\documentclass[10pt,twocolumn,letterpaper]{article}

\usepackage[pagenumbers]{cvpr} 
\usepackage{amsfonts}       
\usepackage{amsmath,amssymb}
\usepackage{multirow} 
\usepackage{makecell}
\usepackage{bm}
\usepackage{url}
\usepackage{cuted}
\usepackage[dvipsnames]{xcolor}

\usepackage{amsmath,amsfonts,bm}

\def\eqref#1{equation~\ref{#1}}

\def\1{\bm{1}}

\DeclareMathAlphabet{\mathsfit}{\encodingdefault}{\sfdefault}{m}{sl}
\SetMathAlphabet{\mathsfit}{bold}{\encodingdefault}{\sfdefault}{bx}{n}

\usepackage{gensymb}
\definecolor{cvprblue}{rgb}{0.21,0.49,0.74}
\usepackage[pagebackref,breaklinks,colorlinks,citecolor=cvprblue]{hyperref}

\def\paperID{4592} 
\def\confName{CVPR}
\def\confYear{2024}

\title{Repaint123: Fast and High-quality One Image to 3D Generation with Progressive Controllable 2D Repainting}
\author{%
    Junwu Zhang$^{1}$\thanks{Equal contribution.} \quad 
    Zhenyu Tang$^{1}$\footnotemark[1] \quad
    Yatian Pang$^{{1},{3}}$ \quad
    Xinhua Cheng$^{1}$ \quad
    Peng Jin$^{1}$ \quad \\
    Yida Wei$^{4}$ \quad
    Munan Ning$^{{1},{2}}$ \quad
    Li Yuan$^{{1},{2}}$\thanks{Corresponding author.} \\
    $^{1}$Peking University 
    $^{2}$ Pengcheng Laboratory
    $^{3}$ National University of Singapore\\
    $^{4}$ Wuhan University
}

\begin{document}
\maketitle

\begin{strip}
    \centering
    \vspace{-5em}
    \centering
    \includegraphics[width=\textwidth]{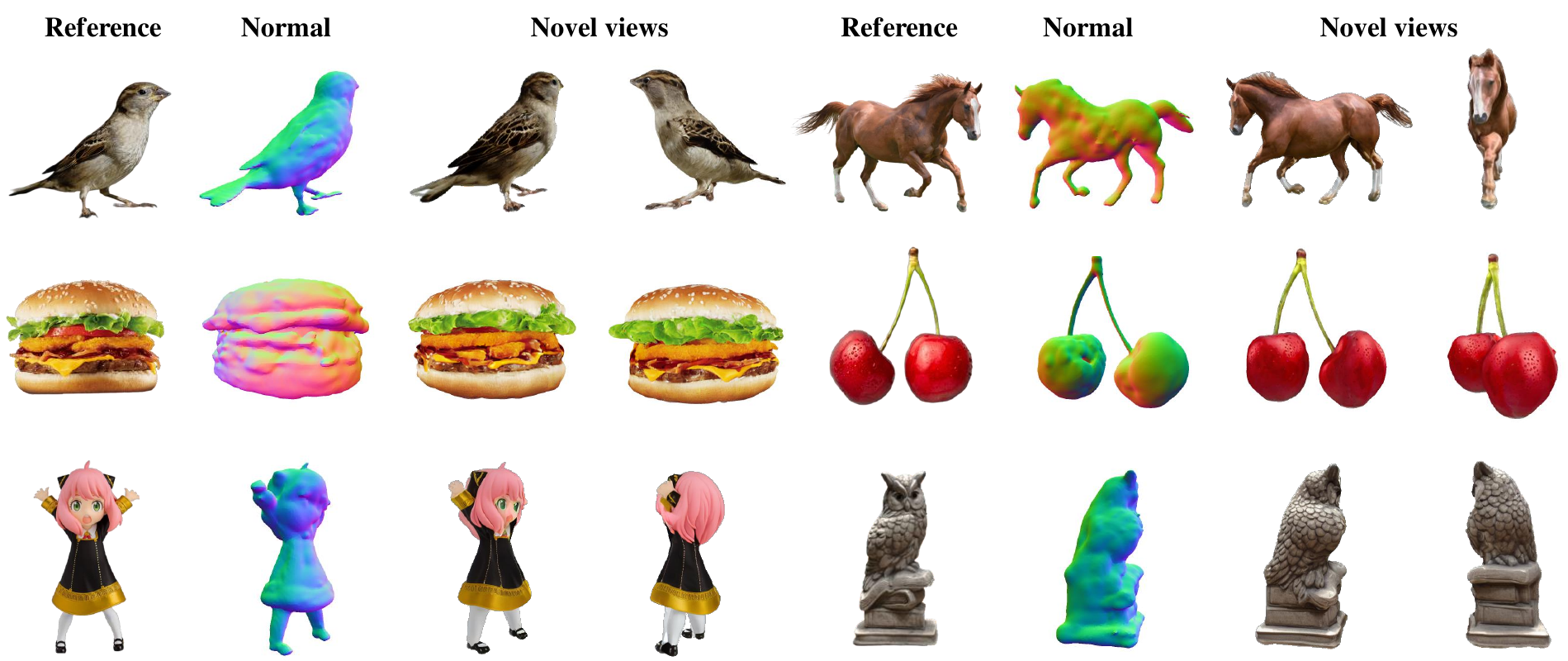}
    \vspace{-2em}
    \captionof{figure}{\textit{Repaint123} generates high-quality 3D content with detailed texture in only \textbf{2 minutes} from a single image. Repaint123 adopts Gaussian Splatting in the coarse stage, and then utilize a 2D controllable diffusion model with repainting stategy to generate view-consistent high-quality images. This allows for fast and high-quality refinement of the extracted mesh texture through simple MSE loss.
    }
    \label{fig:teaser}
\end{strip}

\begin{abstract}

Recent one image to 3D generation methods commonly adopt Score Distillation Sampling (SDS). Despite the impressive results, there are multiple deficiencies including multi-view inconsistency, over-saturated and over-smoothed textures, as well as the slow generation speed. To address these deficiencies, we present Repaint123 to alleviate multi-view bias as well as texture degradation and speed up the generation process. The core idea is to combine the powerful image generation capability of the 2D diffusion model and the texture alignment ability of the repainting strategy for generating high-quality multi-view images with consistency. We further propose visibility-aware adaptive repainting strength for overlap regions to enhance the generated image quality in the repainting process. The generated high-quality and multi-view consistent images enable the use of simple Mean Square Error (MSE) loss for fast 3D content generation. We conduct extensive experiments and show that our method has a superior ability to generate high-quality 3D content with multi-view consistency and fine textures in \textbf{2 minutes} from scratch. Our project page is available at \url{https://pku-yuangroup.github.io/repaint123/}.


\end{abstract}    
\section{Introduction}
\label{sec:intro}

\begin{figure}
    \centering
    \includegraphics[width=0.48\textwidth]{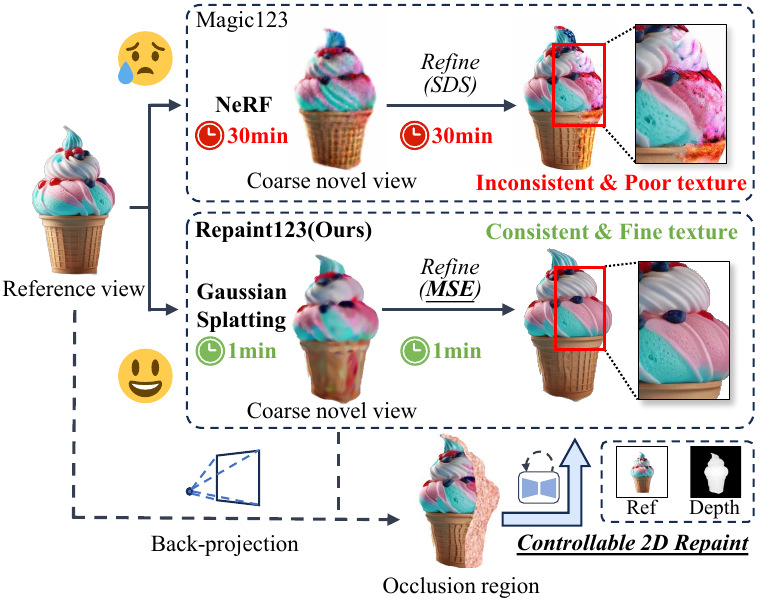}
    \caption{Motivation of our proposed pipeline. Current methods adopt SDS loss, resulting in inconsistent and poor texture. Our idea is to combine the powerful image generation capability of the controllable 2D diffusion model and the texture alignment ability of the repainting strategy for generating high-quality multi-view consistent images. The repainted images enable simple MSE loss for fast 3D content generation.}
    \label{fig:motivation}
\end{figure}

Generating 3D content from one given reference image plays a key role at the intersection of computer vision and computer graphics~\cite{liu2023zero123,liu2023syncdreamer,jun2023shap,nichol2022point,qian2023magic123,dou2023tore}, serving as a pivotal conduit for innovative applications across fields including robotics, virtual reality, and augmented reality. Nonetheless, this task is quite challenging since it is expected to generate high-quality 3D content with multi-view consistency and fine textures in a short period of time. 

Recent studies~\cite{liu2023zero123,lin2023magic3d,melas2023realfusion,tang2023make} utilize diffusion models~\cite{diffusion1,diffusion2}, which have notably advanced image generation techniques, to guide the 3D generation process given one reference image. Generally, the learnable 3D representation such as NeRF is rendered into multi-view images, which then are distilled by rich prior knowledge from diffusion models via Score Distillation Sampling (SDS)~\cite{dreamfusion}. However, SDS may have conflicts with 3D representation optimization~\cite{huang2023dreamtime}, leading to multi-view bias and texture degradation. Despite their impressive results, multiple deficiencies including multi-view inconsistency as well as over-saturated color and over-smoothed textures are widely acknowledged. Moreover,
SDS is very time-consuming as a large number of optimization steps are required. 


To address these deficiencies mentioned above, we propose a novel method called Repaint123 to alleviate multi-view bias as well as texture degradation and speed up the generation process. Our core idea is shown in Figure \ref{fig:motivation} refine stage. We combine the powerful image generation capability of the 2D diffusion model and the alignment ability of the repainting strategy for generating high-quality multi-view images with consistency, which enables using simple Mean Square Error (MSE) loss for fast 3D representation optimization. Specifically, our method adopts a two-stage optimization strategy. The first stage follows DreamGaussian~\cite{tang2023dreamgaussian} to obtain a coarse 3D model in 1 minute. In the refining stage, we first aim to generate multi-view consistent images. For proximal multi-view consistency, we utilize the diffusion model to repaint the texture of occlusion (unobserved) regions by referencing neighboring visible textures. To mitigate accumulated view bias and ensure long-term view consistency, we adopt a mutual self-attention strategy to query correlated textures from the reference view. As for enhancing the generated image quality, we use a pre-trained 2D diffusion model with the reference image as an image prompt to perform classifier-free guidance. We further improve the generated image quality by applying adaptive repainting strengths for the overlap region, based on the visibility from previous views. As a result, with high-quality and multi-view consistent images, we can generate 3D content from these sparse views extremely fast using simple MSE loss.


We conduct extensive one image to 3D generation experiments on multiple datasets and show that our method is able to generate a high-quality 3D object with multi-view consistency and fine textures in about \textbf{2 minutes} from scratch. Compared to state-of-the-art techniques, we represent a major step towards high-quality 3D generation, significantly improving multi-view consistency, texture quality, and generation speed.

Our contributions can be summarized as follows:
\begin{itemize}

\item Repaint123 comprehensively considers the controllable repainting process for image-to-3d generation, preserving both the proximal view and long-term view consistency. 

\item We also propose to enhance the generated view quality by adopting visibility-aware adaptive repainting strengths for the overlap regions.

\item  Through a comprehensive series of experiments, we show that our method consistently demonstrates high-quality 3D content generation ability in \textbf{2 minutes} from scratch.

\end{itemize}

\section{Related Works}
\subsection{Diffusion Models for 3D Generation}
The recent notable achievements in 2D diffusion models~\cite{diffusion1,diffusion2} have brought about exciting prospects for generating 3D objects. Pioneering studies~\cite{dreamfusion, wang2023sjc} have introduced the concept of distilling a 2D text-to-image generation model for the purpose of generating 3D shapes. Subsequent works\cite{chen2023fantasia3d,wang2023prolificdreamer,seo2023ditto,yu2023points,lin2023magic3d,seo2023let,tsalicoglou2023textmesh,zhu2023hifa,huang2023dreamtime,armandpour2023re,wu2023hd,chen2023it3d,tang2023make,melas2023realfusion,qian2023magic123,xu2022neurallift,raj2023dreambooth3d,shen2023anything,cheng2023progressive3d, yu2023hifi} have adopted a similar per-shape optimization approach, building upon these initial works. Nevertheless, the majority of these techniques consistently experience low efficiency and multi-face issues. In contrast to a previous study HiFi-123~\cite{yu2023hifi} that employed similar inversion and attention injection techniques for image-to-3D generation, our approach differs in the selection of diffusion model and incorporation of depth prior. We utilize stable diffusion with ControlNet, introducing depth prior as an additional condition for simplicity and flexibility across various other conditions. In comparison, HiFi-123 employs a depth-based diffusion model (stable-diffusion-2-depth) concatenating depth latent with original latent for more precise geometry control. Meanwhile, we also differ in many other aspects, like the use of repainting strategy, optimization with MSE loss, and Gaussian Splatting representation. 

Recently, some works~\cite{liu2023one, szymanowicz2023viewset, liu2023syncdreamer, long2023wonder3d} extend 2D diffusion models from single-view images to multi-view images to generate multi-view images for reconstruction, while these methods usually suffer from low-quality textures as the multi-view diffusion models are trained on limited and synthesized data.

\subsection{Controllable Image Synthesis}
One of the most significant challenges in the field of image generation has been controllability. Many works have been done recently to increase the controllability of generated images.
ControlNet~\cite{zhang2023controlnet} and T2I-adapter~\cite{mou2023t2i} attempt to control the creation of images by utilizing data from different modalities. Some optimization-based methods~\cite{ruiz2023dreambooth, mokady2023null} learn new parameters or fine-tune the diffusion model in order to control the generation process. Other methods~\cite{ye2023ip-adapter, cao_2023_masactrl} leverage the attention layer to introduce information from other images for gaining better control. 

\subsection{3D Representations}
Neural Radiance Fields (NeRF)~\citep{mildenhall2020nerf}, as a volumetric rendering method, has gained popularity for its ability to enable 3D optimization~\citep{barron2022mipnerf360,li2023neuralangelo,chen2022mobilenerf,hedman2021snerg, Chan2022}  under 2D supervision, while NeRF optimization can be time-consuming. Numerous efforts~\citep{mueller2022instant, yu_and_fridovichkeil2021plenoxels} for spatial pruning have been dedicated to accelerating the training process of NeRF on the reconstruction setting. however, they fail in the generation setting of Nerf. Recently, 3D Gaussian splatting~\cite{kerbl2023gaussiansplatting, chen2023text, yi2023gaussiandreamer, tang2023dreamgaussian}   has emerged as an alternative 3D representation to NeRF and has shown remarkable advancements in terms of both quality and speed, offering a promising avenue. 


\section{Preliminary}

\begin{figure*}
    \centering
    \includegraphics[width=0.95\textwidth]{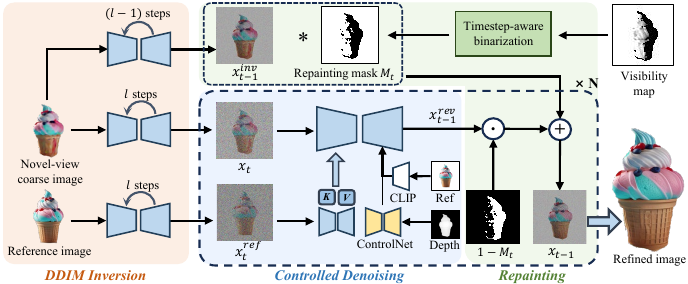}
    \caption{Controllable repainting scheme. Our scheme employs DDIM Inversion~\cite{song2020ddim} to generate deterministic noisy latent from coarse images, which are then refined via a diffusion model controlled by depth-guided geometry, reference image semantics, and attention-driven reference texture. We binarize the visibility map into an overlap mask by the timestep-aware binarization operation. Overlap regions are selectively repainted during each denoising step, leading to the high-quality refined novel-view image.}
    \label{fig:framework}
\end{figure*}

\subsection{DDIM Inversion}
DDIM~\cite{song2020ddim} transforms random noise $\bm{x}_T$ into clean data $\bm{x}_0$ over a series of time steps, by using the deterministic DDIM sampling in the reverse process, i.e., $\bm{x}_{t-1} = (\alpha_{t-1}/\alpha_t)(\bm{x}_t-\sigma_t\epsilon_{\phi})+\sigma_{t-1}\epsilon_{\phi}$. On the contrary, DDIM inversion progressively converts clean data to a noisy state $\bm{x}_T$, i.e., $\bm{x}_{t} = (\alpha_{t}/\alpha_{t-1})(\bm{x}_{t-1}-\sigma_{t-1}\epsilon_{\phi})+\sigma_{t}\epsilon_{\phi}$, here $\epsilon_{\phi}$ is the predicted noise by the UNet. This method retains the quality of the data being rebuilt while greatly speeding up the process by skipping many intermediate diffusion steps.
\subsection{3D Gaussian Splatting}
Gaussian Splatting~\cite{kerbl2023gaussiansplatting} presents a novel method for synthesizing new views and reconstructing 3D scenes, achieving real-time speed. Unlike NeRF, Gaussian Splatting uses a set of anisotropic 3D Gaussians defined by their locations, covariances, colors, and opacities to represent the scene.
To compute the color of each pixel $\mathbf{p}$ in the image, it utilizes a typical neural point-based rendering~\cite{kopanas2021point, kopanas2022neural}, The rendering process is as follows:
\begin{equation}
\begin{split}
        C(\mathbf{p}&) =\sum_{i \in \mathcal{N}} c_i \alpha_i \prod_{j=1}^{i-1}\left(1-\alpha_j\right), \quad \\ \text{where,  } & \alpha_i=o_ie^{-\frac{1}{2}(\mathbf{p}-\mu_i)^T \Sigma_i^{-1}(\mathbf{p}-\mu_i)},
\end{split}
\end{equation}
where $c_i$, $o_i$, $\mu_i$, and $\Sigma_i$ represent the color, opacity, position, and covariance of the $i$-th Gaussian respectively, and $\mathcal{N}$ denotes the number of the related Gaussians.
\section{Method}

In this section, we introduce our two-stage framework for fast and high-quality 3D generation from one image, as illustrated in Figure~\ref{fig:architecture}. In the coarse stage, we adopt 3D Gaussian Splatting as the representation following DreamGaussian~\cite{tang2023dreamgaussian} to learn a coarse geometry and texture optimized by SDS loss. In the refining stage, we convert the coarse model to mesh representation and propose a progressive, controllable repainting scheme for texture refinement. First, we obtain the view-consistency images for novel views by progressively repainting the invisible regions relative to previously optimized views with geometry control and the guidance from reference image(see Section \ref{sec41}). 
Then, we employ image prompts for classifier-free guidance and design an adaptive repainting strategy for further enhancing the generation quality in the overlap regions (see Section \ref{sec42}). Finally, with the generated view-consistent high-quality images, we utilize simple MSE loss for fast 3D content generation.(see Section \ref{sec43}).  

\definecolor{red}{RGB}{197, 90, 17}
\definecolor{blue}{RGB}{46, 117, 182}
\definecolor{yellow}{RGB}{255, 198, 71}
\definecolor{green}{RGB}{84, 130, 53}



\subsection{Multi-view Consistent Images Generation}
\label{sec41}
Achieving high-quality image-to-3D generation is a challenging task because it necessitates pixel-level alignment in overlap regions while maintaining semantic-level and texture-level consistency between reference view and novel views. To achieve this, our key insight is to progressively repaint the occlusions with the reference textures. Specifically, we first delineate the overlaps and occlusions between the reference-view image and a neighboring novel-view image. Inspired by HiFi-123~\cite{yu2023hifi}, we invert the coarse novel-view image to deterministic intermediate noised latents by DDIM Inversion~\cite{song2020ddim} and then transfer reference textures through reference attention feature injection~\cite{cao_2023_masactrl}. The inversion preserves coarse 3D consistent color information in occlusions while the attention injection replenishes consistent high-frequency details. 
Subsequently, we iteratively denoise and blend the noised latent using inverted latents for neighbor harmony and pixel-level alignment in overlaps. 
Finally, we bidirectionally rotate the camera and progressively apply this repainting process from the reference view to all views. By doing this, we can seamlessly repaint occlusions with both short-term consistency (overlaps alignment and neighbor harmony) and long-term consistency (back-view consistency of semantics and textures).

\textbf{Obtaining Occlusion Mask.}
To get the occlusion mask $M_n$ in the novel view with the rendered image $I_n$ and depth map $D_n$, given a repainted reference view with $I_r$ and $D_r$, we first back-project the 2D pixels in the view $V_r$ into 3D points $P_r$ by scaling camera rays of $V_r$ with depth values $D_r$. Then, we render a depth map $D'_n$ from $P_r$ and the target perspective $V_n$. Regions with dissimilar depth values between the two novel-view depth maps ($D_n$ and $D'_n$) are occlusion regions in occlusion mask $M_n$.

\textbf{Performing DDIM Inversion.}
As shown in the red part of Figure~\ref{fig:framework}, to utilize the 3D-consistent coarse color and maintain the textures in overlap regions, we perform DDIM inversion on the novel-view image $I$ to get the intermediate deterministic latents $x^{\text {inv}}$. With the inverted latents, we can denoise reversely to reconstruct the input image faithfully.

\textbf{Repainting the Occlusions with Depth Prior.}
As shown in Figure~\ref{fig:framework}, with the inverted latents, we can replace the overlap parts in the denoised latents during each denoising step to enforce the overlapped regions unchanged while harmonizing the occlusion regions:
\begin{equation}
    x_{t-1} = x_{t-1}^{\text {inv}} \odot (1-M) + x_{t-1}^{\text {rev}} \odot M,
\end{equation}
where $x_{t-1}^{\text {rev}} \sim \mathcal{N}\left(\mu_{\phi}\left(x_{t}, t\right), \Sigma_{\phi}\left(x_{t}, t\right)\right)$ is the denoised latent of timestep $t$. Besides, We employ ControlNet~\cite{zhang2023controlnet} to impose additional geometric constraints from coarse depth maps to ensure the geometric consistency of images.

\textbf{Injecting Reference Textures.}
To mitigate the cumulative texture bias at the back view, we incorporate a mutual self-attention mechanism~\cite{cao_2023_masactrl} that injects reference attention features into the novel-view repainting process during each denoising step. By replacing the novel-view content features (Key features $K_t$ and Value features $V_t$) with reference-view attention features ($K_r$ and $V_r$), the novel-view image features can directly query the high-quality reference features by:
\begin{equation}
    \operatorname{Attention}(Q_t, K_r, V_r)=\operatorname{Softmax}\left(\frac{Q_t K_r^{T}}{\sqrt{d}}\right) V_r,
\end{equation}
where $Q_t$ is the novel-view query features projected from the spatial features. This enhances texture details transfer and improves the consistency of the novel-view image.

\begin{figure*}
    \centering
    \includegraphics[width=0.95\textwidth]{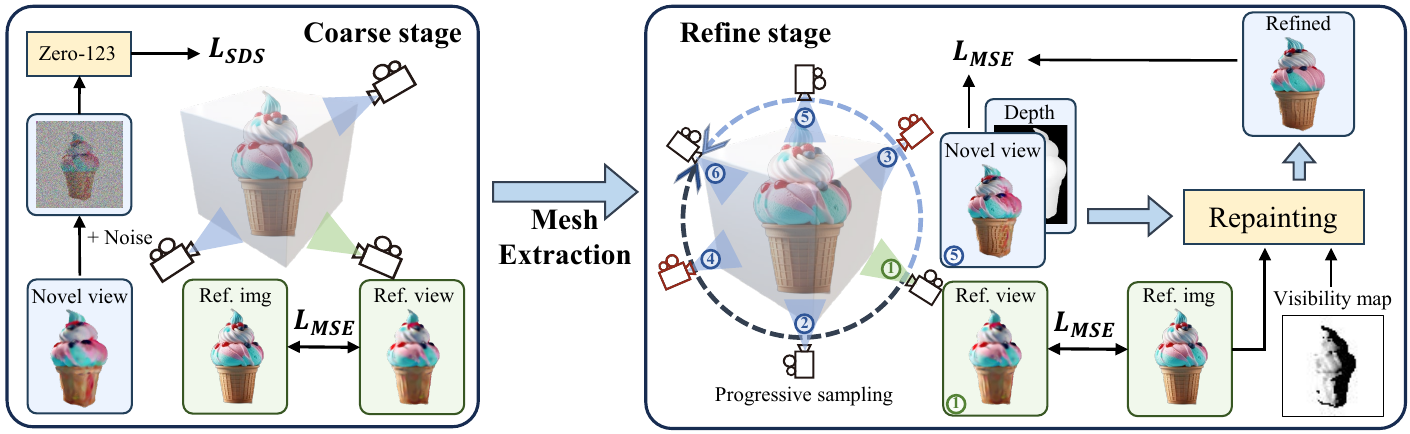}
    \caption{Image-to-3D generation pipeline. In the coarse stage, we adopt Gaussian Splatting representation optimized by SDS loss at the novel view. In the fine stage, we export Mesh representation and bidirectionally and progressively sample novel views for controllable progressive repainting. The novel-view refined images will compute MSE loss with the input novel-view image for efficient generation. Cameras in red are bidirectional neighbor cameras for obtaining the visibility map.}
    \label{fig:architecture}
\end{figure*}

\begin{figure}
    \centering
    \includegraphics[width=0.475\textwidth]{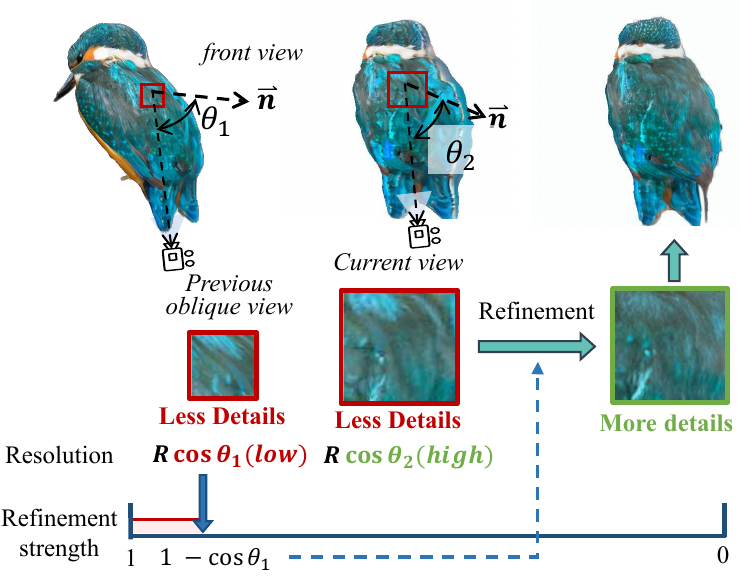}
    \caption{Relation between camera view and refinement strength. The areas in the red box are the same regions from different views.}
    \label{fig:softmask}
\end{figure}

\textbf{Repainting 360\degree  Progressively.}
As shown in Figure~\ref{fig:architecture}, we progressively sample new camera views that alternate between clockwise and counterclockwise increments.
To ensure consistency in the junction of the two directions, our approach selects the nearest camera views from each of the two directions (as shown in the red cameras in Figure~\ref{fig:architecture}) to compute occlusion masks and repaint the texture in the novel view. 
\begin{equation}
M = M_{r} \cap M_{r'},
\end{equation}
where  $M_{r}$ and $M_{r'}$ are the occlusion mask obtained from two reference views, respectively and $M$ is the final mask.

\subsection{Image Quality Enhancement}
\label{sec42}
Despite the progressive repainting process sustaining both short-term and long-term consistency, the accumulated texture degradation over the incremented angles can result in deteriorated multi-view image quality. We discover that the degradation is from both overlap and occlusion regions. For overlap regions, as shown in Figure~\ref{fig:softmask}, when the previous view is an oblique view, it leads to a low-resolution update on the texture maps, resulting in high distortion when rendering from the front view. Therefore, we propose a visibility-aware adaptive repainting process to refine the overlap regions with different strengths based on the previous best viewing angle on these regions. For occlusion regions, they achieve limited quality due to the absence of text prompts to perform classifier-free guidance~\cite{ho2022classifier}, which is essential to diffusion models for high-quality image generation. To improve overall quality, we adopt a CLIP~\cite{clip} encoder (as shown in Figure~\ref{fig:framework}) to encode and project the reference image to image prompts for guidance.

\textbf{Visibility-aware Adaptive Repainting.}
Optimal refinement strength for the overlap regions is crucial, as excessive strength produces unfaithful results while insufficient strength limits quality improvement. 
To select the proper refinement strength, we associate the denoising strength with the visibility map $V$ (similar to the concept of trimap~\cite{richardson2023texture}). As explained in detail in Appendix~\ref{sec:vis}, the visibility map is obtained based on the normal maps (i.e., the $cos \theta$ between the normal vectors of the viewed fragments and the camera view directions) in the current view and previous views. 
For occlusion regions, we set the values in $V$ to $0$.
For overlap regions where the current camera view provides a worse rendering angle compared to previous camera views, we set the values in $V$ to $1$, indicating these regions do not require refinement. For the remaining areas in $V$, we set the values to $cos \theta^*$, which is the largest $cos \theta$ among all previous views and indicates the best visibility during the previous optimization process. 
 In contrast to prior approaches~\cite{richardson2023texture, wang2023breathing} employing fixed denoising strength for all refined fragments, our work introduces a timestep-aware binarization strategy to adaptively repaint the overlap regions based on the visibility map for the faithfulness-realism trade-off. Specifically, as shown in Figure~\ref{fig:softmask}, we view repainting as a process similar to super-resolution that replenishes detailed information. According to the Orthographic Projection Theorem,  which asserts that the projected resolution of a fragment is directly proportional to $cos \theta$, we can assume that the repainting strength is equal to ($1-cos \theta^*$). 
 Therefore, we can binarize the soft visibility map to the hard repainting mask based on the current timestep during each denoising step, denoted by the green box ``Timestep-aware binarization`` in Figure~\ref{fig:framework} and visualized in Figure~\ref{fig:binar}:
\begin{equation}
M_t^{i,j} = \begin{cases}
1, & \text{if  $V^{i,j} > 1 - t/T$} \\
0, & \text{else,} 
\end{cases}
\end{equation}
where $M_t$ is the adaptive repainting mask, $i$, and $j$ are the 2D position of the fragment in visibility map $V$, and $T$ is the total number of timesteps of the diffusion model.




\textbf{Projecting Reference Image to Prompts.}
For image conditioning, previous image-to-3D methods usually utilize textual inversion~\cite{gal2022textualinversion}, which is extremely slow (several hours) for optimization and provides limited texture information due to limited number of learned tokens. Other tuning techniques, such as Dreambooth~\cite{ruiz2023dreambooth}, require prolonged optimization and tend to overfit the reference view. Besides, vision-language ambiguity is a common issue when extracting text from the caption model. To tackle these issues, as shown in Figure~\ref{fig:framework}, we adopt IP-Adapter~\cite{ye2023ip-adapter} to encode and project the reference image into the image prompt of 16 tokens that are fed into an effective and lightweight cross-attention adapter in the pre-trained text-to-image diffusion models. This provides visual conditions for diffusion models to perform classifier-free guidance for enhanced quality.

\subsection{Fast and High-quality 3D Generation}
\label{sec43}
In the coarse stage, we adopt 3D Gaussian Splatting~\cite{kerbl2023gaussiansplatting} with SDS optimization for fast generation. In the fine stage, with the controllable progressive repainting process above, we can generate view-consistent high-quality images for efficient high-quality 3D generation. The refined images are then used to directly optimize the texture through a pixel-wise MSE loss:




\begin{equation}
\mathcal{L}_{\mathrm{MSE}}=||I^{\mathrm{fine}}-I||_2^2,
\end{equation}
where $I^{\mathrm{fine}}$ represents the refined images obtained from controllable repainting and $I$ represent the rendered images from 3D. The MSE loss is fast to compute and deterministic to optimize, resulting in fast refinement.

\section{Experiment}

\begin{table*}[!t]
\centering

\resizebox{\textwidth}{!}{%
\begin{tabular}{c|c|cccc|cc}
\toprule
\multirow{2}{*}{\textbf{Dataset}} & \multirow{2}{*}{\textbf{Metrics \textbackslash \ Methods} }& \multicolumn{4}{c|}{\textbf{NeRF-based}} & \multicolumn{2}{c}{\textbf{Gaussian-Splatting-based}} \\ 
 &  & RealFusion & Make-it-3D & Zero-123-XL* & Magic123 & DreamGaussian & \textbf{Repaint123} \\ 
\midrule
\multirow{4}{*}{\textbf{RealFusion15}} & CLIP-Similarity$\uparrow$ & 0.71 & 0.81 & 0.83 & 0.82 & 0.77 & \textbf{0.85} \\
 & Context-Dis$\downarrow$ & 2.20 &1.82  &1.59  & 1.64  &1.61  &\textbf{1.55}  \\
 & PSNR$\uparrow$ &19.24  &16.56  &19.56  &\textbf{19.68}  &18.94  & \underline{19.00} \\
 & LPIPS$\downarrow$ &0.194 &0.177  &0.108  &0.107  &0.111  &  \textbf{0.101}\\ 
\midrule
\multirow{4}{*}{\textbf{Test-alpha}} & CLIP-Similarity$\uparrow$ &0.68  &0.76  &0.84  &0.84  &0.79  & \textbf{0.88} \\
 & Context-Dis$\downarrow$ &2.20  &1.73  & 1.52  &1.57  &1.62  &\textbf{1.50 } \\
 & PSNR$\uparrow$ & 22.91 &17.21  &24.39  &\textbf{24.69}  & 22.33 &\underline{22.38}  \\
 & LPIPS$\downarrow$ &0.105  &0.237  & 0.050  &\textbf{0.046}  &0.057  & \underline{0.048} \\ 
\midrule
 & Optimization time &20min  &1h  &30min  &1h (+2h)  &\textbf{2min}  &\textbf{2 min}  \\ 
\bottomrule
\end{tabular}
}
\caption{We show quantitative results in terms of CLIP-Similarity$\uparrow$ / Contextual-Distance$\downarrow$ / PSNR$\uparrow$ / LPIPS$\downarrow$. The results are shown on the RealFusion15 and test-alpha datasets, while \textbf{bold} reflects the best for all methods and the \textbf{underline} represents the best for Gaussian-Splatting-based methods. * indicates that Zero123-XL adds a mesh fine-tuning stage to further improve quality. The time required by textual inversion is indicated in parentheses.}
\label{table:image-to-3D}
\end{table*}

\subsection{Implementation Details}
In our experiment, 
we follow DreamGaussian~\cite{tang2023dreamgaussian} to adopt 3D Gaussian Splatting~\cite{kerbl2023gaussiansplatting} representation at the coarse stage. We also explore NeRF as an alternative to Gaussian Splatting for the coarse stage, with results detailed in the Appendix. For all results of our method, we use the same hyperparameters. We progressively increment the viewpoints by 40 degrees, and opt to invert rendered images over 30 steps.
Stable diffusion 1.5 is adopted for all experimented methods.

\begin{table}[t]
\centering

\resizebox{0.48\textwidth}{!}{
\begin{tabular}{c|cccc}
\toprule
 \textbf{Method\textbackslash \ Metric}& \makecell[c]{CLIP}$\uparrow$ & \makecell[c]{Contextual}$\downarrow$ & PSNR$\uparrow$ & LPIPS$\downarrow$ \\
\midrule
\makecell[c]{Coarse} &0.71  &1.78  & 21.17 & 0.133 \\
\makecell[c]{\textbf{repaint}} &0.71  &1.62  & 22.41 &0.049 \\
\makecell[c]{\textbf{+mutual attention}} &0.78  &1.56  &22.42  &0.048  \\
\textbf{+image prompt} &0.84  &1.52  &22.40  &  0.048\\
\textbf{+adaptive (Full)} & 0.88  &1.50  & 22.38 &  0.048\\
\bottomrule
\end{tabular}
}
\caption{Quantitative ablation study on Test-alpha dataset.}
\label{table:ablation}
\end{table}
\begin{figure*}
    \centering
    \includegraphics[width=0.94\textwidth]{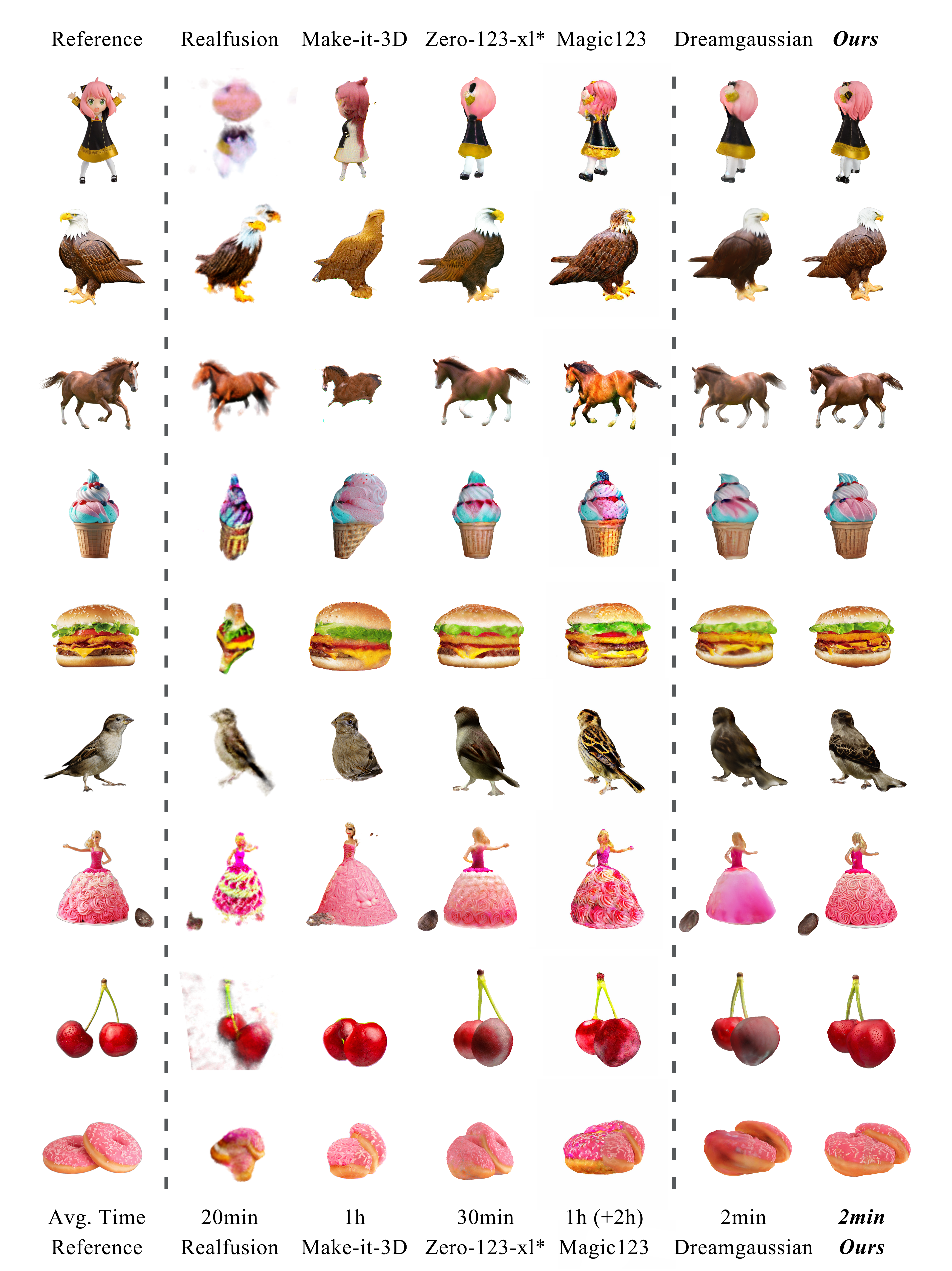}
    \caption{Qualitative comparisons on image-to-3D generation. Zoom in for texture details. }
    \label{fig:result}
\end{figure*}
\subsection{Baselines}
We adopt RealFusion~\cite{melas2023realfusion}, Make-It-3D~\cite{tang2023make}, and Zero123-XL~\cite{liu2023zero123}, Magic123~\cite{qian2023magic123} as our NeRF-based baselines and DreamGaussian~\cite{tang2023dreamgaussian} as our Gaussian-Splatting-based baseline. RealFusion presents a single-stage algorithm for NeRF generation leveraging an MSE loss for the reference view along with a 2D SDS loss for novel views. Make-It-3D is a two-stage approach that shares similar objectives with RealFusion but employs a point cloud representation for refinement at the second stage. Zero123 enables the synthesis of novel views conditioned on images without the need for training data, achieving remarkable quality in generating 3D content when combined with SDS. Integrating Zero123 and RealFusion, Magic123 incorporates a 2D SDS loss with Zero123 for consistent geometry and adopts DMTet~\cite{shen2021dmtet} representation at the second stage. DreamGaussian integrates 3D Gaussian Splatting into 3D generation and greatly improves the speed. For Zero123-XL, we adopt the implementation~\cite{stable-dreamfusion}, For other works, we use their officially released code for evaluation.

\subsection{Evaluation Protocol}

\textbf{Datasets.} Based on previous research, we utilized the Realfusion15 dataset~\cite{melas2023realfusion} and test-alpha dataset  collected by Make-It-3D~\cite{tang2023make}, which comprises many common things. 

\textbf{Evaluation metrics.} An effective 3D generation approach should closely resemble the reference view, and maintain consistency of semantics and textures with the reference when observed from new views. Therefore, to evaluate the overall quality of the generated 3D object, we choose the following metrics from two aspects: 
1) PSNR and LPIPS~\cite{zhang2018lpips}, which measure pixel-level and perceptual generation quality respectively at the reference view;
2) CLIP similarity~\cite{clip} and contextual distance~\cite{mechrez2018contextual}, which assess the similarity of semantics and textures respectively between the novel perspective and the reference view.


\subsection{Comparisons}

\textbf{Quantitative Comparisons.} 
As shown in Table~\ref{table:image-to-3D}, we evaluate the quality of generated 3D objects across various methods. Our method achieves superior 3D consistency in generating 3D objects, as evidenced by best performance from CLIP-similarity and contextual distance metrics. Although our method achieves better reference-view reconstruction results than DreamGaussian, there is a gap compared with Nerf-based approaches, which we attribute to the immaturity of current Gaussian-Splatting-based methods. Compared with Nerf-based methods for the optimization time, our approach reaches a significant acceleration of over 10 times and simultaneously achieves high quality.


\textbf{Qualitative Comparisons.}
Figure~\ref{fig:result} displays the qualitative comparison results between our method and the baseline, while Figure~\ref{fig:teaser} shows multiple novel-view images generated by our methods. 
Repaint123 achieves the best visual results in terms of texture consistency and generation quality as opposed to other methods. From the visual comparison in Figure~\ref{fig:result}, we discover that DreamGaussian and  Zero123-XL usually result in over-smooth textures, lowering the quality of the 3D object generation. Magic123 often produces inconsistent oversaturated colors in invisible areas. Realfusion and Make-It-3D fail to generate full geometry and consistent textures. This demonstrates Repaint123's superiority over the current state of the art and its capacity to generate high-quality 3D objects in about 2 minutes.

\begin{figure}
    \centering
    \includegraphics[width=0.475\textwidth]{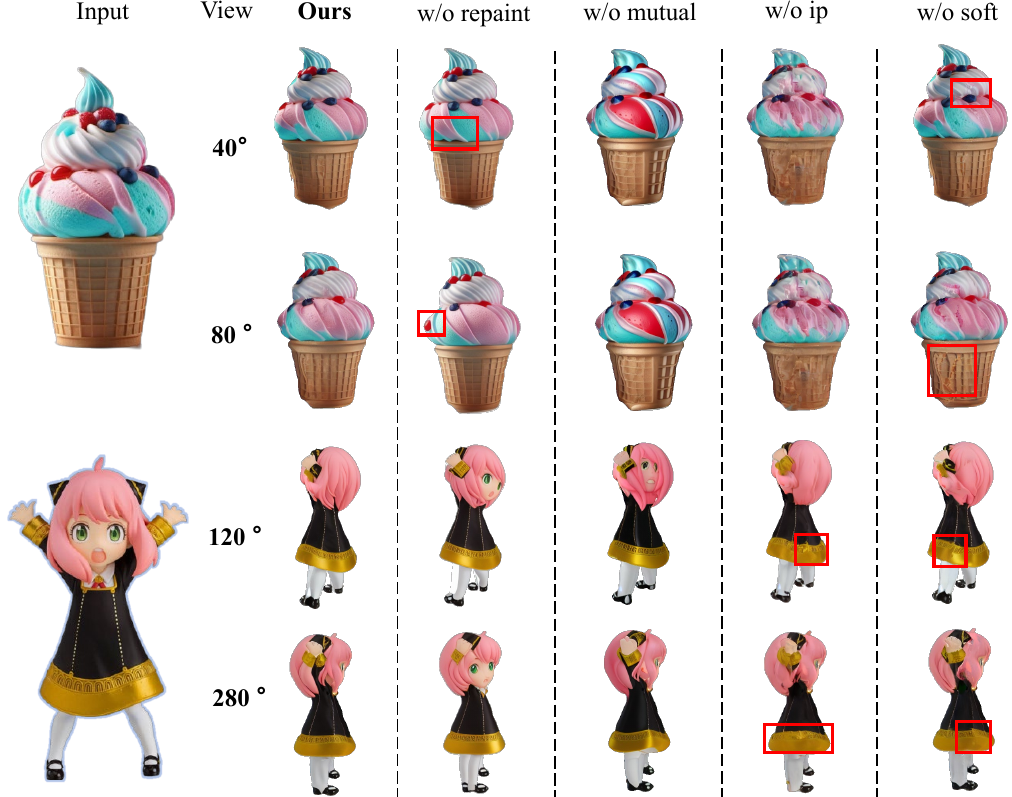}
    \caption{Qualitative ablation study. Red boxes show artifacts.}
    \label{fig:ab}
\end{figure}

\subsection{Ablation and Analysis}
In this section, we further conduct both qualitative and quantitative ablation studies (as shown in Figure~\ref{fig:ab} and Table~\ref{table:ablation}) to demonstrate the effectiveness of our designs. Furthermore, we analyze the angular interval during repainting.

\textbf{Effectiveness of Progressive Repainting.} As shown in Figure~\ref{fig:ab}, there are noticeable multi-face issues (anya) and inconsistencies in the overlap regions (ice cream). These inconsistencies are from the absence of constraints on overlap regions and can introduce conflicts and quality degradation of the final 3D generation. 


\textbf{Effectiveness of Mutual Attention.} In Table~\ref{table:ablation}, we can see that mutual attention can significantly improve the multi-view consistency and fine-grained texture consistency compared to vanilla repainting. As shown in Figure~\ref{fig:ab}, the synthesized images without mutual attention strategy maintain the semantics but fails to transfer detailed textures from the reference image.

\textbf{Effectiveness of Image Prompt.} As shown in Table~\ref{table:ablation}, image prompt can further improve the multi-view consistency. As shown in Figure~\ref{fig:ab} and Table~\ref{table:ablation}, without an image prompt for classifier-free guidance, the multi-view images fail to generate detailed realistic and consistent textures. 

\textbf{Effectiveness of Adaptive Repainting.} As shown in Figure~\ref{fig:ab}, without the adaptive repainting mask, the oblique regions in the previous view will lead to artifacts when facing these regions due to previous low-resolution updates. Table~\ref{table:ablation} also demonstrates the effectiveness as both CLIP similarity and Contextual distance improve significantly.



\begin{table}[t]
\centering

\resizebox{0.48\textwidth}{!}{
\begin{tabular}{c|cccc}
\toprule
  \textbf{Angle\textbackslash \ Metric}& \makecell[c]{CLIP}$\uparrow$ & \makecell[c]{Contextual}$\downarrow$ & PSNR$\uparrow$ & LPIPS$\downarrow$ \\
\midrule
\makecell[c]{20\degree} &0.873  &1.504  &22.35  & 0.051 \\
\makecell[c]{40\degree} & 0.881
 &1.506  &22.38  &0.048  \\
\makecell[c]{60\degree} &0.888  &1.497  &22.27  &0.050  \\
\makecell[c]{80\degree} &0.885 
  & 1.487
  &22.26  &0.051  \\
\bottomrule
\end{tabular}
}
\caption{Effects of the chosen angle of neighboring viewpoints in Repaint123 on Test-alpha dataset.}
\label{table:Angle ablation}
\end{table}

\begin{figure}[!t]
    \centering
    \includegraphics[width=0.48\textwidth]{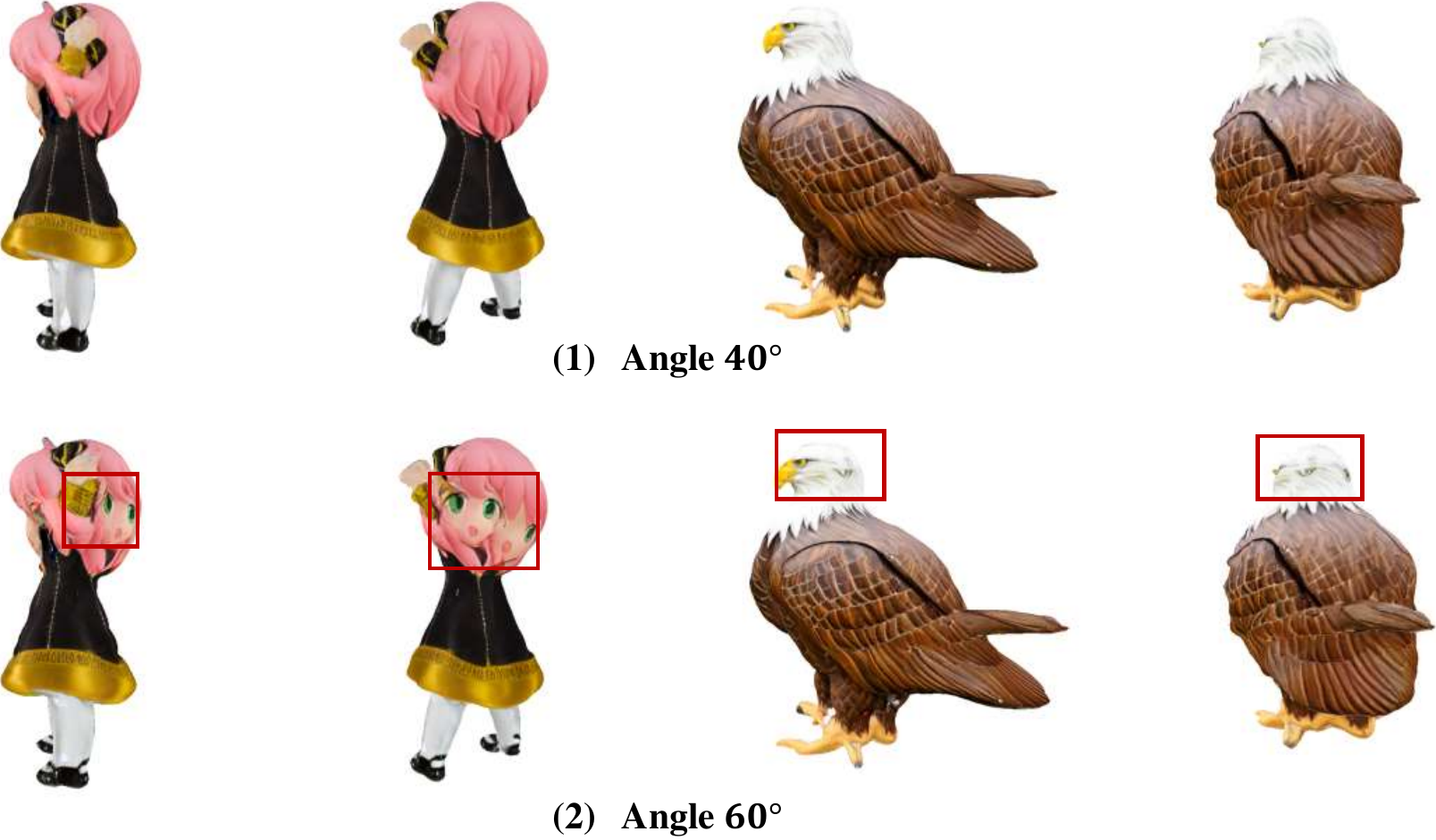}
    \caption{Visual comparison when choosing 40\degree and 60\degree as the angle interval. The red box shows the resulting multi-face issues.}
    \label{fig:angle}
\end{figure}



\textbf{Analysis of Angle Interval.}  
We study the effect of using different angle intervals on the performance of Repaint123 in Table~\ref{table:Angle ablation}. The table demonstrates that the metrics achieve their peak performance when the angle is set to 60 degrees.
Nonetheless, Figure~\ref{fig:angle} illustrates that there is a reduced overlapping area when choosing 60 degrees as the angle interval, which consequently increases the likelihood of encountering a multi-head problem during the optimization process. Thus, we ultimately choose 40 degrees as the ideal angle interval for the optimization process.



\section{Discussion}
While Gaussian Splatting is fast, due to the lack of technological maturity for generation tasks and mesh extraction, it may exhibit geometry artifacts, such as  holes, and achieve inferior results compared to NeRF-based methods. These issues are expected to be resolved with its development.


\section{Conclusion}
This work presents Repaint123 for generating high-quality 3D content from a single image in about 2 minutes.  By leveraging progressive controllable repaint, our approach overcomes the limitations of existing studies and achieves state-of-the-art results in terms of both texture quality and multi-view consistency, paving the way for future progress in one image 3D content generation. 
Furthermore, we validate the effectiveness of our proposed method through a comprehensive set of experiments.


{
    \small
    \bibliographystyle{ieeenat_fullname}
    \bibliography{main}
}
\clearpage
\clearpage
\maketitlesupplementary



\section{Visibility Map and Repainting Strength}
\label{sec:vis}

This section delineates our proposed visibility map and its relation to the repainting strength in detail and visualization.

\begin{figure}
    \centering
    \includegraphics[width=0.48\textwidth]{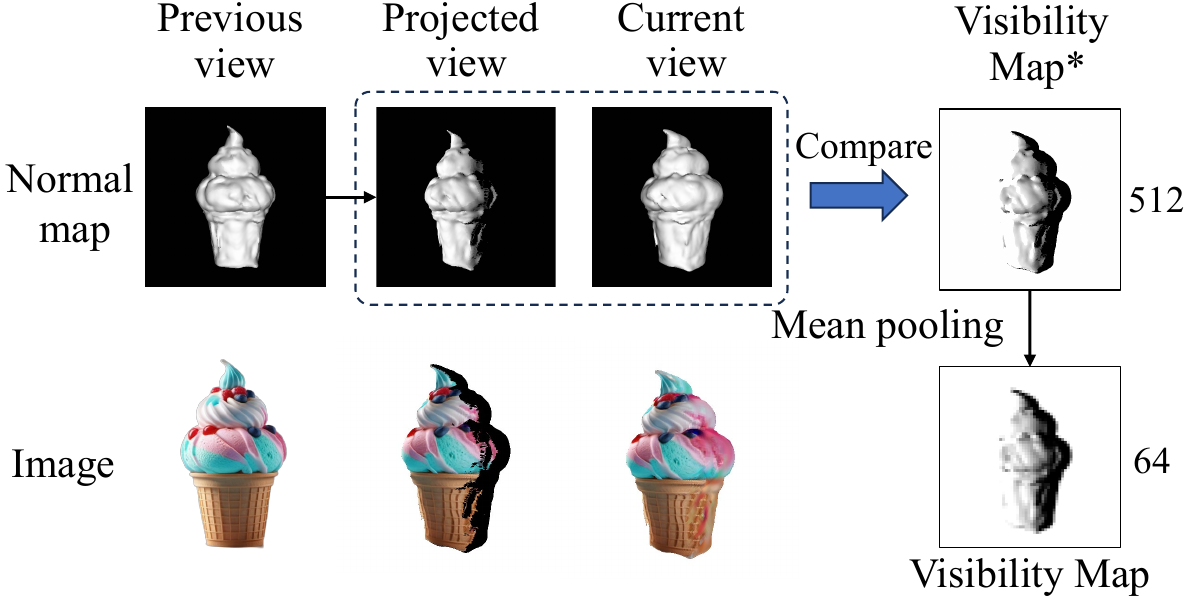}
    \caption{Visibility map creation process. The value in the normal map represents the visibility. White parts of the visibility map are less visible regions in the current view compared to the previous views, while non-white parts are more visible regions with the value of previous visibility.}
    \label{fig:vis}
\end{figure}

\textbf{Obtaining Visibility Map.}
Figure~\ref{fig:vis} shows the process of transforming the novel-view normal map to the visibility map based on the previous neighbor-view normal map. We first conduct a back-projection of the preceding normal map into 3D points, subsequently rendering a normal map from the novel view based on these 3D points, i.e., the normal map in the projected view as shown in Figure~\ref{fig:vis}. Comparing novel-view normal maps with the projected novel-view normal maps yields a high-resolution visibility map, assigning projected normal map values to areas with improved visibility in the novel view (non-white parts of visibility map in Figure~\ref{fig:vis}) for further refinement and a value of 1 to other regions (white parts of visibility map in Figure~\ref{fig:vis}) for preservation. The final visibility map is obtained by downsampling from 512x512 to 64x64 resolution, facilitating subsequent repainting mask generation in the latent space.

\begin{figure}
    \centering
\includegraphics[width=0.48\textwidth]{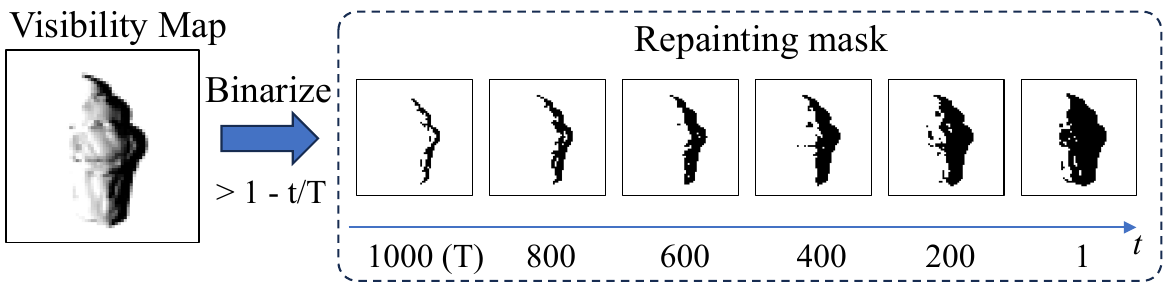}
    \caption{Timestep-aware binarization. Black areas represent repainting regions, and white areas denote preservation regions.}
    \label{fig:binar}
\end{figure}

\textbf{Timestep-aware Binarization.}
As shown in Figure~\ref{fig:binar}, we visualize our proposed timestep-aware binarization process to transform the visibility map into the timestep-dependent repainting mask. Based on the proportional relation between visibility and repainting strength elucidated in Figure~3 in the main paper, the repainting region (black areas in Figure~\ref{fig:binar}) can be obtained by selecting areas in the visibility map with a visibility value not exceeding $1-t/T$, where $T$ represents the maximum timestep during training (typically 1000), and $t$ denotes the current repainting timestep. As illustrated in Figure~\ref{fig:binar},  decreasing denoising timesteps enlarges repainting regions, indicating a progressive refinement according to prior visibility.

\section{Evaluation on Multi-view Dataset}
We adopt the Google Scanned Object (GSO) dataset~\cite{downs2022google} and use 10 objects for multi-view evaluation of the generated 3D objects with 3D ground truth. As shown in Table~\ref{table: GSO}, the results indicate that our method is capable of
generating high-quality 3D contents with multi-view consistency compared with the strong baselines.

\section{Evaluation of NeRF-based Repaint123}
\begin{figure*}[!t]
    \centering
\includegraphics[width=0.97\textwidth]{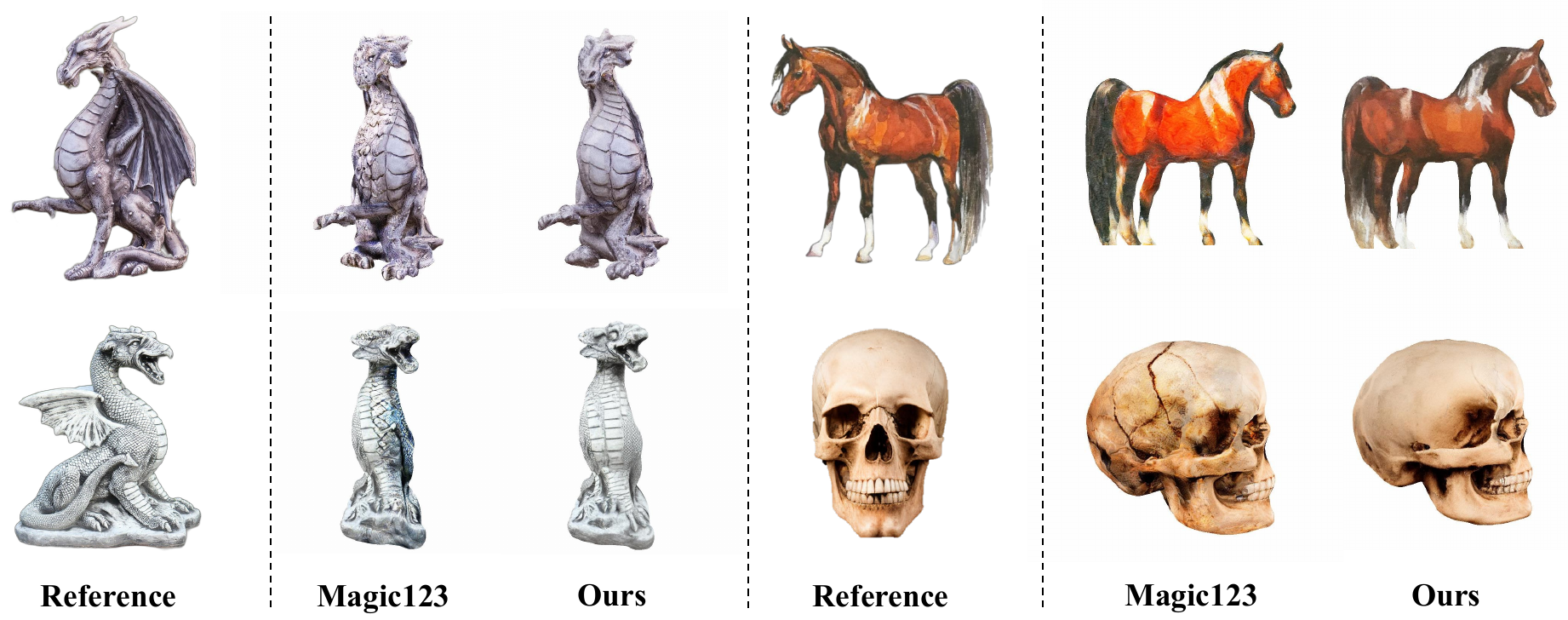}
    \caption{Visual comparison between our NeRF-based method and Magic123.}
    \label{fig:Magic123-our}
\end{figure*}

As our repainting approach is plug-and-play for the refinement stage, we can change the representation in the coarse stage from Gaussian Splatting to NeRF. As presented in Table~\ref{table:supple-image-to-3D} and Figure~\ref{fig:Magic123-our}, the generated 3D objects can be significantly improved by using our repainting method.

\begin{table}[t]
\centering

\resizebox{0.48\textwidth}{!}{
\begin{tabular}{c|ccccc}
\toprule
 \textbf{Method\textbackslash \ Metric}& PSNR$\uparrow$ & SSIM$\uparrow$& LPIPS$\downarrow$&CLIP$\uparrow$ &Contextual$\downarrow$\\
\midrule
\makecell[c]{Syncdreamer} &13.201 &0.784 &0.322& 0.612&1.686\\
\makecell[c]{Magic123} &14.985 &0.803  &0.244 &0.767&1.376 \\
\makecell[c]{Zero-123-XL} &15.118 &0.813 &0.229 &0.761& 1.334\\
\makecell[c]{DreamGaussian} &15.391  &0.814  & 0.237 &0.736&1.407 \\
\textbf{Repaint123} &\textbf{15.393} &\textbf{0.814} &\textbf{0.214}&\textbf{0.812}&\textbf{1.319}\\
\bottomrule
\end{tabular}
}
\caption{ Multi-view quantitative comparison with image-to-3D generation baselines on GSO dataset.}
\label{table: GSO}
\end{table}

\begin{table*}[!t]
\centering

\resizebox{0.92\textwidth}{!}{%
\begin{tabular}{c|c|cc|cc}
\toprule
\multirow{2}{*}{\textbf{Dataset}} & \multirow{2}{*}{\textbf{Metrics \textbackslash \ Methods} }& \multicolumn{2}{c|}{\textbf{NeRF-based}} & \multicolumn{2}{c}{\textbf{Gaussian-Splatting-based}} \\ 
 &   & Magic123  & \textbf{Ours*} & DreamGaussian & \textbf{Ours} \\ 
\midrule
\multirow{4}{*}{\textbf{RealFusion15}} & CLIP-Similarity$\uparrow$ & 0.82 &\textbf{0.85} & 0.77 & \textbf{0.85} \\
 & Context-Dis$\downarrow$ & 1.64  &\textbf{1.57} &1.61  &\textbf{1.55}  \\
 & PSNR$\uparrow$ &19.68  & \textbf{20.27} &18.94  & \textbf{19.00} \\
 & LPIPS$\downarrow$ &0.107  &\textbf{0.096} &0.111  &  \textbf{0.101}\\ 
\midrule
\multirow{4}{*}{\textbf{Test-alpha}} & CLIP-Similarity$\uparrow$ &0.84  &\textbf{0.88} &0.79  & \textbf{0.88} \\
 & Context-Dis$\downarrow$  &1.57 &\textbf{1.46} &1.62  &\textbf{1.50 } \\
 & PSNR$\uparrow$ &24.69 &\textbf{24.91}  & 22.33 &\textbf{22.38}  \\
 & LPIPS$\downarrow$ &0.046 &\textbf{0.036} &0.057  & \textbf{0.048} \\ 
\bottomrule
\end{tabular}
}
\caption{We show quantitative results in terms of CLIP-Similarity$\uparrow$ / Contextual-Distance$\downarrow$ / PSNR$\uparrow$ / LPIPS$\downarrow$. The results are shown on the RealFusion15 and test-alpha datasets, while \textbf{bold} reflects the best for Nerf-based and Gaussian-Splatting-based methods respectively. * indicates that we adopt NeRF representation for the coarse stage.}
\label{table:supple-image-to-3D}
\end{table*}

\begin{table}[t]
\centering

\resizebox{0.48\textwidth}{!}{
\begin{tabular}{c|cccc}
\toprule
 \textbf{Prompt\textbackslash \ Metric}& PSNR$\uparrow$ & LPIPS$\downarrow$ & CLIP$\uparrow$ & Contextual$\downarrow$ \\
\midrule
\makecell[c]{None} &19.02 &0.102 &0.79&1.60 \\
\makecell[c]{Text} &19.00 &0.102 &0.83 &1.58\\
\makecell[c]{Textual Inversion} &19.01 & 0.101 &0.84 &1.57 \\
\textbf{Image} &19.00 & \textbf{0.101}&\textbf{0.85}&\textbf{1.55}\\
\bottomrule
\end{tabular}
}
\caption{Ablation on RealFusion15 dataset under various prompt conditions. Image prompt achieves superior performance.}
\label{table:prompt}
\end{table}

\section{Ablation Study on Prompt}
In this section, we conduct ablations on various prompts, including image prompt, text prompt, textual inversion, and empty prompt.
As shown in Table~\ref{table:prompt}, prompts significantly enhance both multi-view consistency and quality in comparison to results obtained without prompts. The efficacy stems from the classifier-free guidance. Among various prompts, image prompts demonstrate superior performance, showcasing the superior accuracy and efficiency of visual prompts over text prompts, including time-consuming optimized textual prompts.

\section{More Results}
The videos in the supplementary material show more image-to-3D generation results of our method,  demonstrating our method can produce high-quality 3D contents with consistent appearances.

\end{document}





\input{sec/X_suppl}